# Image-Derived $PM_{10}$ Estimation in Cattle Feedlot Using Machine Learning: Addressing Concentration Ranges Beyond Existing Digital Imaging Methods

Sirapoom Peanusaha[a], Greg B. Ferguson[a], K. Jack Bush[a], Peiyang Li[a,b], Brent W. Auvermann[a,b*]

[a] Texas A&M AgriLife High Plains Research and Extension Center, 3211 Russell Long Blvd., Canyon, TX 79015 United States of America

[b] Texas A&M University, Department of Biological and Agricultural Engineering, College Station, TX 77843 United States of America

* Corresponding author contact: Brent.Auvermann@ag.tamu.edu

## Abstract

Affordable dust monitoring remains a pressing need for the cattle feedlot industry, yet camera-based PM estimation, despite its growing body of research in urban air quality settings, has not been evaluated under the extended concentration ranges characteristic of intensive livestock operations. This study developed an image-based approach using contrast panel features and machine learning to estimate $PM_{10}$ concentrations in a commercial cattle feedlot, where hourly average $PM_{10}$ ranged from 250 to 1,000 μg $m^{-3}$ and instantaneous concentrations reached 5,000 to 20,000 μg $m^{-3}$. Grayscale images were captured during the evening dust peak period, and features including panel contrast, black and white panel pixel values, and overall image brightness were extracted. The model also incorporated recent past values from preceding images and solar zenith angle as predictors. Among the candidate models evaluated, XGBoost achieved the highest predictive performance, with an $R^2$ of 0.792 and a median absolute error of 103 μg $m^{-3}$. Feature importance analysis revealed that (a) panels positioned farthest from the camera contributed most strongly to predictions and (b) that black panel pixel values were more sensitive than white panel values to changes in $PM_{10}$ concentration. Prediction accuracy during the sunset transition, which coincides with the onset of the feedlot evening dust peak, remains an area for further refinement. These findings demonstrate the feasibility of image-based $PM_{10}$ estimation across PM concentration ranges substantially exceeding those reported in prior urban studies and provide practical guidelines for future deployment in feedlot environments.

## 1. Introduction

Cattle production is a cornerstone of U.S. agriculture, with the Central Plains region accounting for a substantial share of national beef output. As of 2022, Texas alone maintained over 12.5 million head of all cattle and calves (encompassing beef, dairy, and other classes) (USDA National Agricultural Statistics Service, 2024), supporting more than 54,000 agricultural jobs and contributing billions of dollars annually to the state economy (Jones et al., 2024). The cattle feeding industry faces increasing scrutiny regarding air emissions from Concentrated Animal Feeding Operations (CAFOs), particularly concerning particulate matter emissions (Bonifacio et al., 2011; Bush et al., 2014; Razote et al., 2008; Sweeten et al., 1988). Particulate matter (PM) with aerodynamic diameter less than 10 micrometers ($PM_{10}$) poses health risks for both humans and animals, causing respiratory infections (Emert et al., 2024; Habib et al., 2024), organic dust toxic syndrome (Poole and Romberger, 2012; Tarlo et al., 2021), and allergies (May et al., 2012). The long-term sustainability of feedlots, especially as the interface with their rural neighbors shrinks, hinges upon success in accurately measuring and reducing feedlot dust emissions (Bonifacio et al., 2012).

Addressing these environmental concerns requires reliable and economically feasible monitoring technologies capable of quantifying PM emissions across the region's numerous feedlot facilities. Currently available instrumentation for PM measurement includes Tapered Element Oscillating Microbalance (TEOM), BETA Attenuation Monitoring (BAM) and optical devices based on light scattering technology (Upadhyay et al., 2008). These instruments are expensive, difficult to calibrate and require constant maintenance. Due to these factors they tend to be beyond the economic reach of feedlot operators and are typically regionally deployed by governmental agencies. This regional deployment results in low spatial resolution for point source PM emissions (Pudasaini et al., 2020). Beyond cost and complexity, these instruments differ in how they sample the surrounding air, using one of two measurement modes: point or path averaged. Point measurement quantifies PM concentration at a single fixed location, as exemplified by the TEOM, but may not adequately represent a spatially dispersed source such as a feedlot. Path-averaged measurement instead determines the mean concentration along a defined path between transmissometer-based equipment, such as a laser or light-based sensor, and its receiver, providing an integrated picture better suited to large, non-uniform sources. This distinction is an important consideration when selecting a monitoring approach for feedlot applications. To aid a feedlot operator in practically addressing PM emissions from their facility, an accurate, reliable and economically feasible alternative PM monitoring system is needed. The system must be readily available, easily deployed and low maintenance.

Lower-cost commercial sensors have been considered as potential alternatives to regulatory-grade instruments. Devices such as the Davis AirLink model 7210 (Davis Instrumentation, Hayward, CA, USA) and the Temtop PMD 351 (Temtop Inc, San Jose, CA, USA) offer reduced upfront costs but are primarily designed for indoor air quality monitoring or periodic spot checks and are not suitable for continuous outdoor monitoring in high PM concentration environments such as in agricultural operations (Rowland, 2024), reflecting a broader challenge in which monitoring technologies developed for urban environments often cannot be directly transferred to agricultural settings (Shaw et al., 2004) Mid-range instruments such as the Met One ES-642 Remote Dust Monitor (Met One Instruments, Grants Pass, OR, USA) and the Thermo Fisher Scientific 1405 TEOM Continuous Ambient Particulate Monitor (Thermo Fisher Scientific, Waltham, MA, USA) are more appropriate for high-concentration outdoor environments but remain cost-prohibitive for widespread adoption at the individual operation level (See Appendix 1). Digital image analysis has emerged as a low-cost alternative for PM concentration estimation, with demonstrated accuracy across a wide range of urban environments (Liu et al., 2016, 2024; Wang et al., 2024; Xu et al., 2019; Yao et al., 2022).

Digital image analysis for PM concentration estimation is a well-studied approach in urban air quality monitoring. The typical study setup involves retrieving scene imagery from widely available RGB images captured by stationary cameras such as traffic or surveillance cameras, paired with PM or AQI concentrations recorded at nearby weather monitoring stations, with machine learning or deep learning models trained to recognize patterns of visual change in the images corresponding to changes in pollution levels. Liu et al. (2024) used traffic camera RGB images captured four times per hour to estimate $PM_{2.5}$ concentrations, with ground truth measurements obtained from a monitoring station located 3 km away, achieving an $R^2$ of approximately 0.98 and RMSE of 0.76 μg $m^{-3}$ for $PM_{2.5}$. Wang et al. (2024) employed Long Short-Term Memory (LSTM) models paired with scene images from online databases to predict $PM_{10}$, $PM_{2.5}$, and AQI levels across three datasets, with reference monitoring stations located between 1.40 and 3.9 km from the image capture locations, reporting $R^2$ values ranging from 0.6 to 0.9 for $PM_{10}$ concentrations.

Despite the demonstrated accuracy of these urban-focused methodologies, significant knowledge gaps remain in applying these techniques to rural agricultural settings, presenting three distinct challenges that motivate the present study.

First, urban studies are generally conducted within relatively low ambient $PM_{10}$ concentration ranges, with reported maximum $PM_{10}$ values typically between 70 and 834 μg $m^{-3}$ and dataset mean concentrations often around 30 to 50 μg $m^{-3}$ (Carretero-Peña et al., 2019; Kow et al., 2022; Wang et al., 2024). In cattle feedlot environments, $PM_{10}$ concentrations can reach well beyond 3,000 μg $m^{-3}$, with hourly average concentrations during evening dust peak periods ranging from 400 to 1,000 μg $m^{-3}$ (Emert et al., 2024; Hiranuma et al., 2011; Peanusaha et al., 2026). To the authors' knowledge, no existing digital image-based study has evaluated the effectiveness of this approach across the extended $PM_{10}$ concentration ranges characteristic of intensive livestock operations.

Second, urban studies commonly rely on the spatial homogeneity of ambient $PM_{10}$ concentrations across city environments, with reference monitoring stations placed as far as 2 to 4 km from the image capture location (Feng et al., 2021; Liu et al., 2024; Wang et al., 2024). This spatial separation may not be appropriate in feedlot settings, where dust concentrations can change dramatically over short distances from the emission source due to the highly localized and dynamic nature of cattle activity, vehicle traffic, and pen management operations (Auvermann et al., 2010; Razote et al., 2006).

Third, urban image-based studies commonly extract features from prominent structural edges such as building outlines to train deep learning models on unstructured scene imagery (Liaw and Chen, 2021; Wang et al., 2024, 2022). These visual cues are absent in open rural agricultural landscapes. Furthermore, urban studies benefit from large volumes of images sourced from publicly available camera networks already installed for other purposes, providing abundant training data for deep learning approaches. In contrast, camera deployment on private feedlot operations requires producer investment, resulting in limited training data that makes unstructured deep learning approaches less practical. These combined challenges highlight the need for a purpose-designed monitoring approach that accounts for the unique optical, spatial, and operational conditions of feedlot environments. Sakirkin et al. (2010) demonstrated the feasibility of using painted contrast panel as a proxy for dust estimation in a feedlot setting, representing the only known prior application of image-based analysis in this context. The present study builds directly upon that foundational work by incorporating machine learning and extending the approach to a substantially broader $PM_{10}$ concentration range with a larger dataset.

The primary goal of this study is to advance the application of digital image analysis for $PM_{10}$ estimation in feedlot environments, building upon prior foundational work to provide producers with an accessible and economically feasible tool for monitoring dust emissions. To achieve this, the study pursues two specific objectives. The first objective is to develop an automated algorithm to extract image-derived features from contrast panel photographs and train machine learning models to predict $PM_{10}$ concentrations referenced against co-located TEOM measurements. The second objective is to characterize the relationships between camera-to-panel distance, ambient lighting conditions, and measurement system configuration, with the aim of establishing practical deployment guidelines that support the translation of this approach into usable monitoring technology for feedlot producers.

# 2. Materials and Methods

## 2.1 Experiment Location

The collaborating feedyard is a typical 2 $km^2$ (500-acre) 50,000 head facility located in the semi-arid Texas Panhandle primarily engaged in finishing yearling heifers and steers for slaughter. Cattle are fed 180-190 days to achieve desired weight before shipping. The climate is typically hot and dry during summer, with peak temperatures reaching 37–40 °C and relative humidity of 15–20%, while winter lows can drop to −6 °C. Wind is typically from the south or southwest at speeds ranging from 16-48 km/h.

## 2.2 Camera and $PM_{10}$ Reference Instrument Deployment

Two co-located Canon EOS 60D digital SLR cameras (Canon, Ota City, Japan) equipped with Canon EF 200mm L II Ultrasonic lenses were used to take photographs of camera targets. Cameras were mounted side by side inside a 91.4 cm x 152.4 cm x 50.8 cm enclosed insulated aluminum cabinet. They were positioned directly in front of two 8.9 cm circular view ports in the face of the cabinet. The view ports were covered on the exterior of the cabinet by square panes of 2.4 mm glass plate mounted in a frame. Cabinet was equipped with heating and cooling unit to mitigate overheating of interior and mitigate fogging/condensation on camera lenses and port cover glass.

Operation of the cameras was accomplished using proprietary Canon software allowing a remote operator to access and set up the cameras to begin taking photographs at a specified time, at a specified interval and to take a specified number of pictures. Raw images from the cameras were RGB images in both .CR2 and .jpg formats. Images were saved on a computer connected to each camera and located in the enclosure with the cameras. The cameras were timed to begin taking photographs at the beginning of the evening dust peak. The cameras were set to take a photograph every 5 minutes and to continue to take photographs for the duration of the dust event. The two downwind TEOM sensors (DW03 and DW04) were positioned at distances of 65 m (DW03) and 200 m (DW04) downwind from the camera. $PM_{10}$ concentrations were recorded by each TEOM sensor as 5-minute average values, consistent with the 5-minute image capture interval of the camera system. The beginning and duration of evening dust peak was derived from PM10 concentration levels recorded by the two downwind TEOMs. Data collection spanned approximately 18 months, from winter 2024 to fall 2025, thereby capturing conditions across all four seasons. Over this period, each camera acquired a total of approximately 3,300 images.

Photo targets consisted of 12 individual target panels, each panel painted with two rectangular targets, one a white rectangle surrounded by a black border, the other a black rectangle surrounded by a white border. The painted targets touched in the middle on each target panel. Target panels were arranged into side-by-side pairs at each of 6 distances from the cameras. The size of the painted targets on each panel increased in size from 23.0 cm x 21.0 cm at the closest target panels (panels #11 and #12) to 119.0 cm x 100.0 cm at the furthest target panels (panels #1 and #2). The size and spacing of each panel pair were designed so that the targets would occupy approximately the same number of pixels in the captured images, regardless of distance. Accordingly, panels positioned farther from the cameras were proportionally larger to compensate for their reduced apparent size in the cameras' field of view. The panels were tilted approximately 6 degrees downward to reduce glare. The two downwind TEOM sensors (DW03 and DW04) were positioned near the approximate centroid of the area-source dust plume from the feedyard, in accordance with the prevailing wind direction.

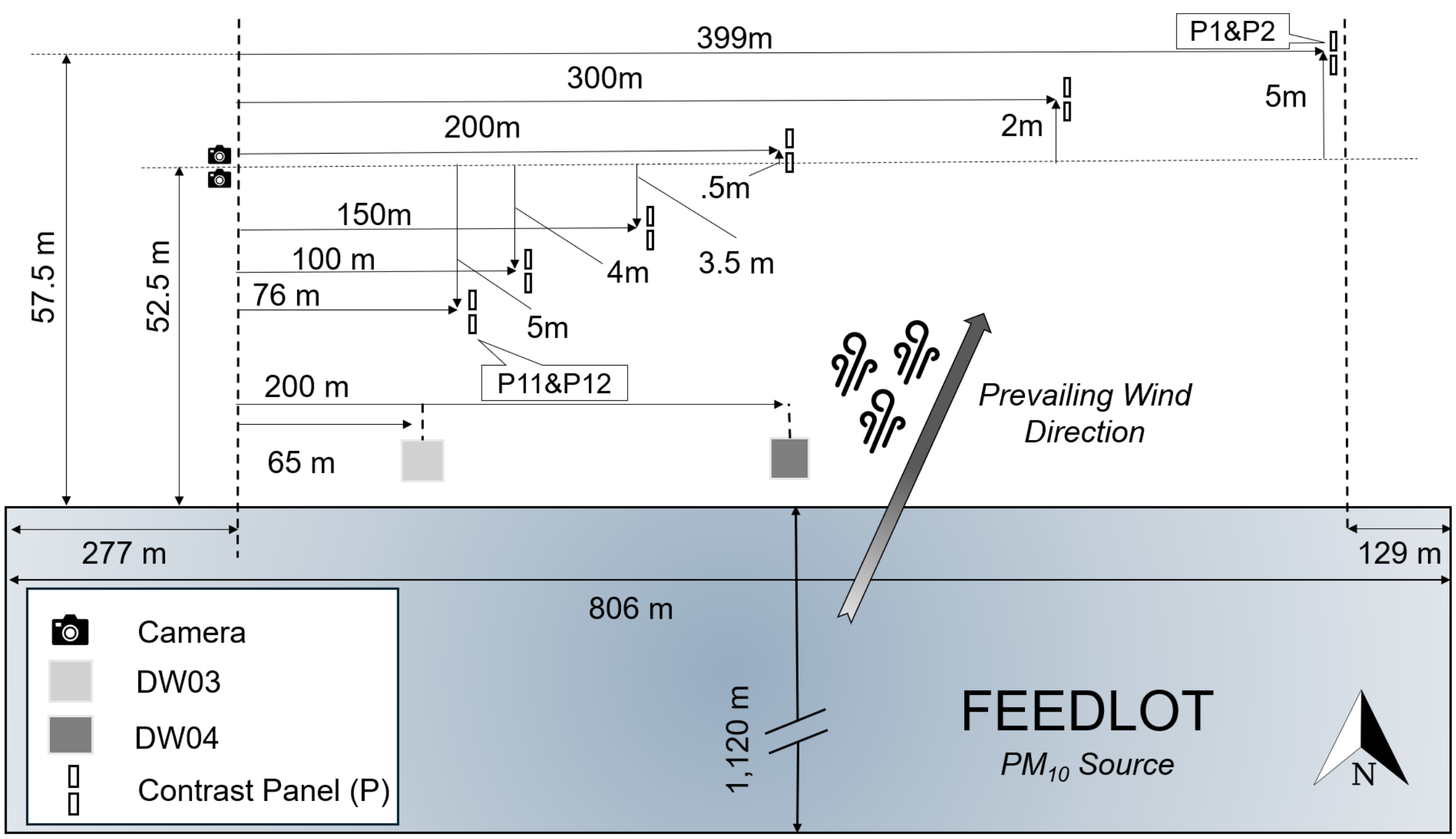

Figure 1. Top-down view of the $PM_{10}$ concentration sensing setup. Digital cameras capture reflected light from six target pairs spaced between 75 m and 400 m downwind of the feedlot. TEOM reference samplers are positioned at 65 m and 200 m to validate image-derived $PM_{10}$ estimates. Panels are numbered sequentially from P1 and P2 (furthest from the camera) to P11 and P12 (closest to the camera). Note: Figure not drawn to scale.

Raw images from cameras were converted from proprietary Canon .CR2 color (RGB) format to a grayscale image in a lossless .tiff format using National Television System Committee (NTSC) standard equation described in Sakirkin et al (2010).

$$I = 0.2989R + 0.5870G + 0.1140B \quad (1)$$

R, G, and B represent red, green, and blue luminosity values and I represents greyscale intensity of each pixel inside the image.

## 2.3 Panel segmentation and value extraction

Following conversion to 8-bit grayscale TIFF format, images were processed for automated panel segmentation. Pixel intensity filtering was applied using thresholds of 0 to 60 for black targets and 195 to 255 for white targets. To suppress noise and consolidate target regions, morphological closing followed by opening (3×3 kernel) was applied to each binary mask. Contiguous regions (blobs) were then extracted from the resulting masks, and only those with dimensions consistent with panel specifications, namely widths of approximately 25 to 60 pixels and heights of approximately 70 to 110 pixels, were retained as candidate targets. The segmentation algorithm subsequently identified valid black and white target pairs meeting the following criteria: (i) centroids aligned vertically within 30 pixels along the y-axis, (ii) black and white targets of comparable size, with differences in both width and height not exceeding 20 pixels, and (iii) horizontal separation between the paired centroids of approximately 20 to 150 pixels along the x-axis. A schematic illustration of these geometric criteria and the associated image coordinate frame is provided in Appendix 3. Finally, duplicate detections were removed by discarding pairs whose centroids fell within 50 pixels of a previously retained pair, keeping the larger of the two. Images with all 12 panels successfully detected were retained as candidate anchor images for the subsequent recovery step.

For images where initial segmentation failed due to suboptimal lighting conditions, an anchor-based approach was implemented. Given the fixed spatial position of panels across the deployment period, despite minor variations in camera angle due to maintenance activities, structural vibration, or wind-induced displacement, panel coordinates from successfully segmented images within the same day were used as initial position estimates. The anchor image was selected as the image with all 12 panels detected and the largest total panel pixel area, representing the clearest detection within that day. A search window spanning twice the panel dimensions (100% of the panel width and height added around the anchored coordinates) was established to identify black or white regions meeting the original segmentation criteria. Detection was attempted on the black sub-panel first, with the white sub-panel used as fallback if the black sub-panel could not be identified. A solidity threshold of 0.70 was applied to reject fragmented or noise-induced detections. Once either sub-panel centroid was located, the full panel boundary was reconstructed geometrically, with the panel center estimated by offsetting one quarter of the panel width laterally from the detected sub-panel centroid. Panel boundaries were then delineated at these refined positions.

Only images where all 12 panels were successfully segmented, either through initial detection or anchor-based recovery, were retained for feature extraction. Contrast values were extracted using a standardized region-of-interest approach. A 10×20-pixel patch was sampled from the approximate center of each panel, and mean pixel intensity was calculated for both black and white panels. Panel contrast was quantified as the difference in mean intensity between paired white and black panels. This process yielded a final dataset of 4,067 labeled images.

## 2.4 Feature Engineering and Machine Learning Model

### 2.4.1 Image-$PM_{10}$ Data Pairing and Image Feature Extraction

Each image was temporally paired with $PM_{10}$ concentrations recorded by both DW03 and DW04. Because image acquisition was initiated manually, capture times did not always align precisely with the top of each hour. For example, a session nominally scheduled for 19:00:00 might begin at 19:01:02. To account for this, each image was matched to the nearest available $PM_{10}$ observation in time, subject to a maximum allowable offset of five minutes. Following the pairing step, a set of radiometric and geometric features was extracted from each image. Panel pixel values for both the black and white reference panels were computed from a 10 × 20-pixel region centered on each panel blob. Consistent with the areal-mean approach described by Sakirkin et al. (2010) and Kwon (2004), the mean pixel value across this region was used as the representative panel value. Two brightness metrics were derived: a whole-image brightness defined as the mean pixel intensity across all image pixels, and a panel brightness defined as the mean pixel value within the bounding box of the white reference panel. Solar geometry was also incorporated as a predictor, given the well-established influence of sun zenith angle on incoming irradiance in remote sensing applications (Jafarbiglu and Pourreza, 2023; Peanusaha, 2025) and its demonstrated relevance in urban digital-image-based PM estimation (Liu et al., 2016). The solar zenith angle for each image was computed using the pvlib library (Holmgren et al., 2018), with inputs comprising the latitude and longitude of the feedlot site and the recorded timestamp of each image. Additionally, image sharpness at each panel location was quantified using the Laplacian variance, computed by applying the Laplacian operator to the grayscale pixel values within each panel bounding box and calculating the variance of the resulting response; higher Laplacian variance indicates a sharper, more in-focus panel region, while lower values indicate blur consistent with optical attenuation by suspended particulate matter. This was implemented using the cv2.Laplacian function from the OpenCV library (Bradski, 2000) applied to each panel crop independently, yielding twelve panel-level Laplacian variance values per image.

### 2.4.2 Machine Learning Model Development

Three machine learning regressors were employed in this study: Random Forest Regressor (RFR) (Breiman, 2001), Extreme Gradient Boosting (XGB) (Chen and Guestrin, 2016), and Quantile Random Forest (QRF) (Meinshausen, 2006). Machine learning approaches were preferred over deep learning for two reasons. First, the training dataset in this study is limited in size, a condition under which machine learning models are known to generalize more reliably than data-hungry deep learning (Chauhan and Singh, 2018; Wang et al., 2021). Second, unlike studies involving surveillance cameras where scene composition and camera placement vary, this study employs a standardized panel positioned at a fixed distance from the camera, producing structured, consistent image features that are well-suited to machine learning methods designed for tabular data.

RFR was selected as an established ensemble method with demonstrated applicability to image-derived features in environmental monitoring contexts. Although prior applications have primarily relied on satellite imagery, RFR has shown strong predictive performance in comparable remote sensing regression tasks (Zheng et al., 2020). XGB was included on the basis of its consistently strong performance across regression benchmarks (Chen and Guestrin, 2016) and its demonstrated potential in image-based air quality estimation in urban smartphone studies (Liu et al., 2026). QRF was selected to address the distributional characteristics of the target variable. The PM10 measurements, particularly from the DW04 sensor, exhibited a right skewed distribution. As QRF estimates the full conditional distribution of the response variable rather than solely the conditional mean, it is better equipped to produce accurate predictions under such distributional conditions.

In addition to per-image features, lag features derived from images captured within a defined time window prior to the target image at t = 0 were incorporated into the feature set. This inclusion was motivated by the argument of Wang et al. (2024) that air pollution is not a static process, as temporal continuity exists in its evolution from one moment to the next. Accordingly, lag features were introduced to provide the model with contextual information regarding the visual appearance of the panel in the period preceding each measurement, enabling the model to account for temporal trends in particulate matter concentration rather than treating each observation in isolation. The optimal lag length was determined empirically by evaluating model performance across lag steps ranging from 1 to 5, with the upper bound constrained by data availability, as each additional lag step requires a corresponding number of preceding images within the same session and increases the proportion of observations that must be discarded from the training and evaluation dataset due to incomplete lag histories.

Both DW03 and DW04 $PM_{10}$ concentrations were predicted simultaneously using a multi-output regression framework. This approach was motivated by evidence that jointly predicting correlated target variables can improve overall accuracy relative to training independent single-output models. This is particularly relevant in the context of co-located air quality sensors, as demonstrated by Kang et al. (2026), who showed improved prediction accuracy when simultaneously estimating multiple co-varying pollutants including $PM_{10}$, $PM_{2.5}$, $NO_2$, $O_3$, $SO_2$, and CO using a multi-output framework, as well as in agricultural modeling applications (Nguyen et al., 2023; Tsakiridis et al., 2020).

## 2.5 Model Evaluation

Model performance was evaluated using five complementary metrics: Root Mean Square Error (RMSE), Normalized Root Mean Square Error (NRMSE), the coefficient of determination ($R^2$),

Median Absolute Error (MedAE), as defined in Equations 2–5, respectively. RMSE quantifies the magnitude of prediction error in the original unit of measurement, making it directly interpretable in the physical context of the target variable. NRMSE normalizes this error relative to the mean of observed values, enabling fair comparison across models operating at different scales or trained on different datasets. $R^2$ measures the proportion of variance in the observed data explained by the model, providing an indication of overall goodness of fit. Because RMSE and NRMSE square the residuals, they assign disproportionate weight to large errors; this is a notable limitation for the present dataset, which contains substantial extreme $PM_{10}$ values arising from intermittent dust events. To provide error estimates that are robust to these extremes, two additional metrics were included. Median Absolute Error (MedAE) is the median of absolute errors and, unlike mean-based metrics, is largely unaffected by a small number of extreme deviations, thereby reflecting the model's typical error magnitude.

To assess the generalizability and robustness of each model, five-fold cross-validation was employed throughout all experiments. The dataset was partitioned into five non-overlapping folds, with each fold serving as the held-out test set in turn while the remaining four folds were used for training. Performance metrics were computed for each fold and subsequently averaged to yield a final estimate that is less susceptible to the influence of any particular data split.

An ablation study was additionally conducted to examine the risk of data leakage introduced by the zenith angle feature. Because solar zenith angle is a deterministic function of time, its inclusion in a forecasting model trained on historical data carries the risk of implicitly encoding lead time information. The ablation study systematically evaluated model performance with and without the zenith angle to isolate its contribution.

$$RMSE = \sqrt{\frac{1}{n}\sum_{i=1}^{n}(\hat{y}_i - y_i)^2} \quad (2)$$

$$NRMSE = \frac{RMSE}{\bar{y}} \times 100 \quad (3)$$

$$R^2 = 1 - \frac{\sum_{i=1}^{n}(y_i - \hat{y}_i)^2}{\sum_{i=1}^{n}(y_i - \bar{y})^2} \quad (4)$$

$$MedAE = median(|y_1 - \hat{y}_1|, |y_2 - \hat{y}_2|, \ldots, |y_n - \hat{y}_n|) \quad (5)$$

# 3. Results

## 3.1 Study Period $PM_{10}$ Levels and Image-Derived Panel Contrast Characteristics

Figure 2 presents the temporal distribution of images captured throughout the study period following the panel extraction process. Image acquisition was concentrated during the evening hours (16:00–21:00 local time) to maximize the dynamic range of $PM_{10}$ concentrations recorded. This sampling strategy aligns with well-documented diurnal patterns of dust emission in feedlot environments, where peak emissions occur during evening hours as solar radiation decreases, temperatures drop, and animal activity intensifies (Marcillo and Auvermann, 2025; Peanusaha et al., 2026). DW04 consistently recorded higher $PM_{10}$ concentrations than DW03 across all hours (Figure 2b). This is likely attributable to the spatial positioning of DW04 relative to the emission source. DW04 lies closer to the plume centerline, situated more directly downwind of the area-source centroid along the dominant wind direction (Appendix 2). Consequently, DW04 intercepts a higher concentration of transported particulate matter than DW03, which is positioned further from the plume core.

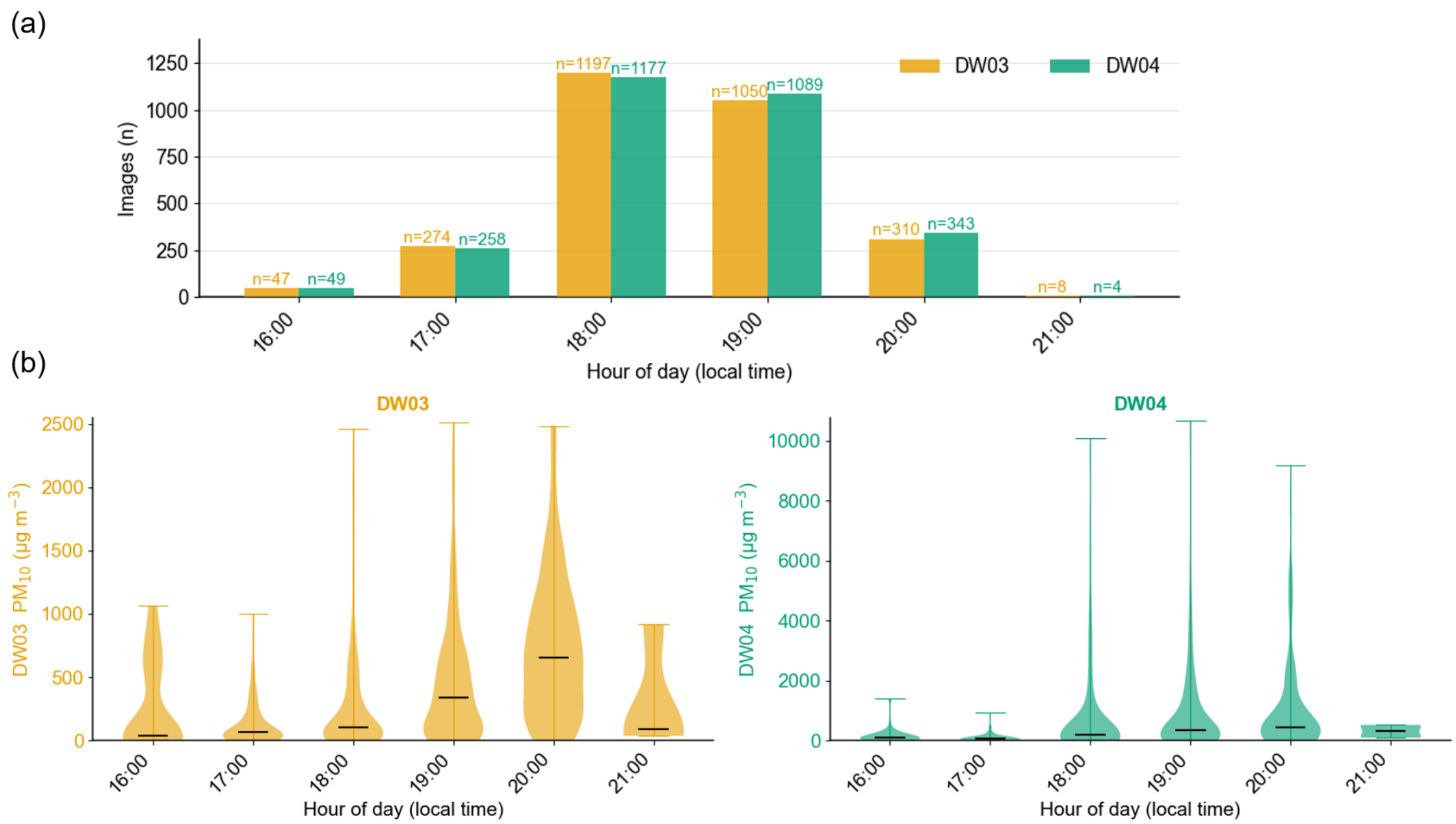

Figure 2. Temporal distribution of measurements from downwind sensors DW03 and DW04 during the study period. (a) Hourly count of images captured at each sensor between 16:00 and 21:00 local time. (b) Violin plots showing the distribution of $PM_{10}$ concentrations (µg $m^{-3}$) recorded at 5-minute intervals for DW03 (left) and DW04 (right) across the same hourly periods.

Figure 3 illustrates the visual difference in panel contrast between low and high dust conditions captured on representative dates. The overlaid values represent contrast, calculated as the white minus black pixel value for each panel pair. Under low dust conditions (Figure 3a), contrast values remain relatively consistent across all distances, ranging from approximately 197 at the nearest panel to 191 at the farthest, indicating minimal atmospheric attenuation. In contrast, under high dust loading (Figure 3b), contrast values decline with increasing distance from the camera, dropping from approximately 193 at the nearest panel to 150 at the furthest panels, visually demonstrating the contrast-reduction principle upon which this study is based. Furthermore, within-pair variation at the same distance increases with distance, reflecting greater spatial heterogeneity in dust concentration as the plume is sampled farther from the camera.

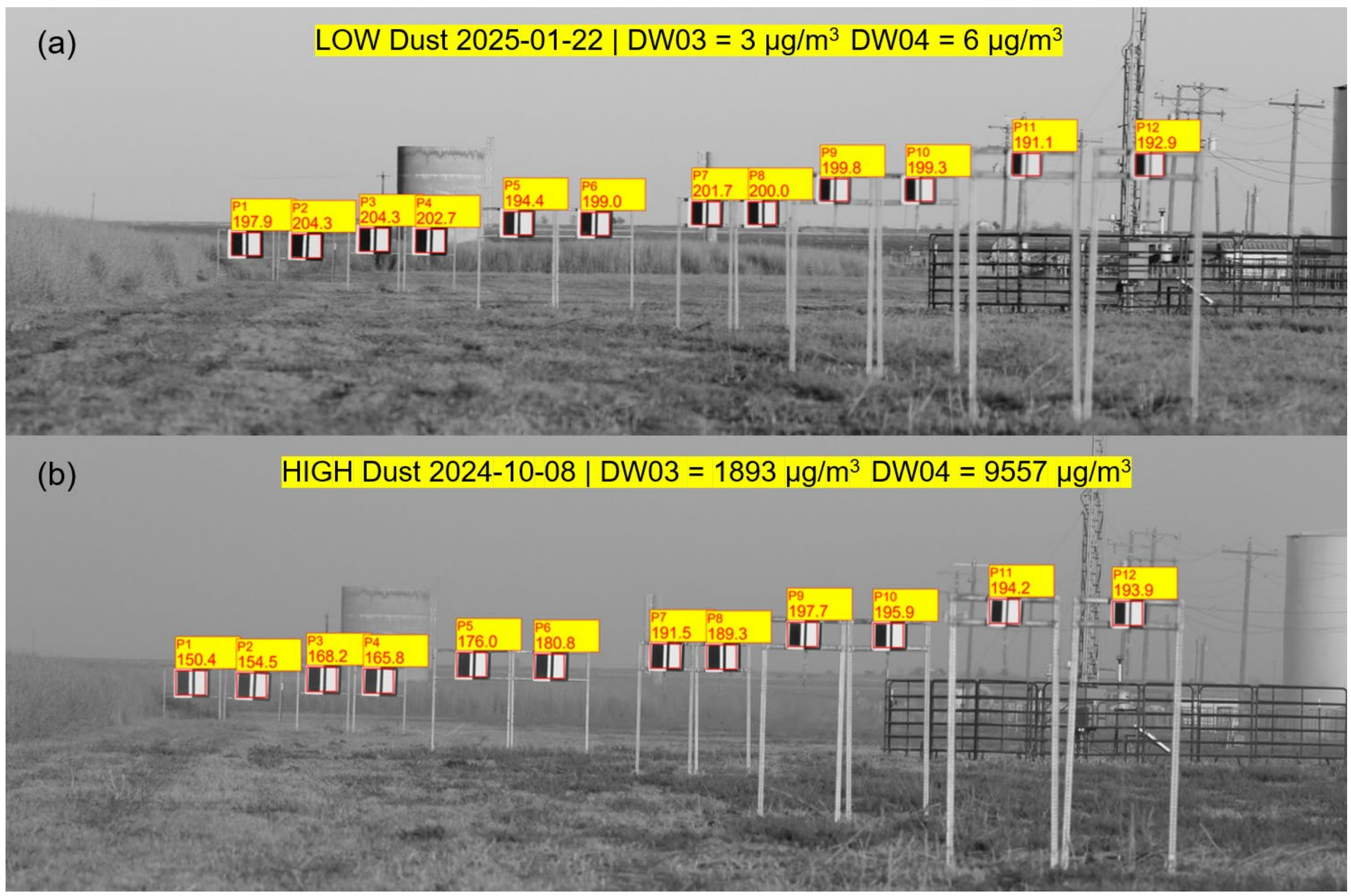

Fig 3. Representative camera images captured under (a) low dust and (b) high dust conditions. Red bounding boxes indicate the detected panel regions, with the overlaid value representing the contrast calculated as the white minus black panel pixel value. Under high dust conditions, contrast values decline noticeably with increasing distance from the camera.

Figure 4 presents the mean panel contrast at each panel distance, grouped by DW03 $PM_{10}$ concentration quartiles. Under low dust conditions (Q1–Q2), contrast remains stable across all distances, with values consistently near 185–200. As $PM_{10}$ concentration increases in Q3 (210–657 µg $m^{-3}$) and Q4 (660–6,613 µg $m^{-3}$), a clear downward trend in contrast emerges with increasing distance, indicating that suspended dust progressively reduces the visual difference between the black and white panels.

Notably, the black and white panels respond differently to rising $PM_{10}$ levels. The mean white panel pixel value remained relatively stable between 200 and 210 across all quartiles, while the mean black panel pixel value increased substantially, rising from approximately 35 in Q1–Q2 to nearly 85 in Q4. This asymmetric response suggests that the black panel is considerably more sensitive to changes in $PM_{10}$ concentration, as airborne dust scatters light onto the darker surface more detectably than onto the already near-saturated white panel. Consequently, the reduction in overall contrast is driven primarily by the brightening of the black panel rather than the darkening of the white panel.

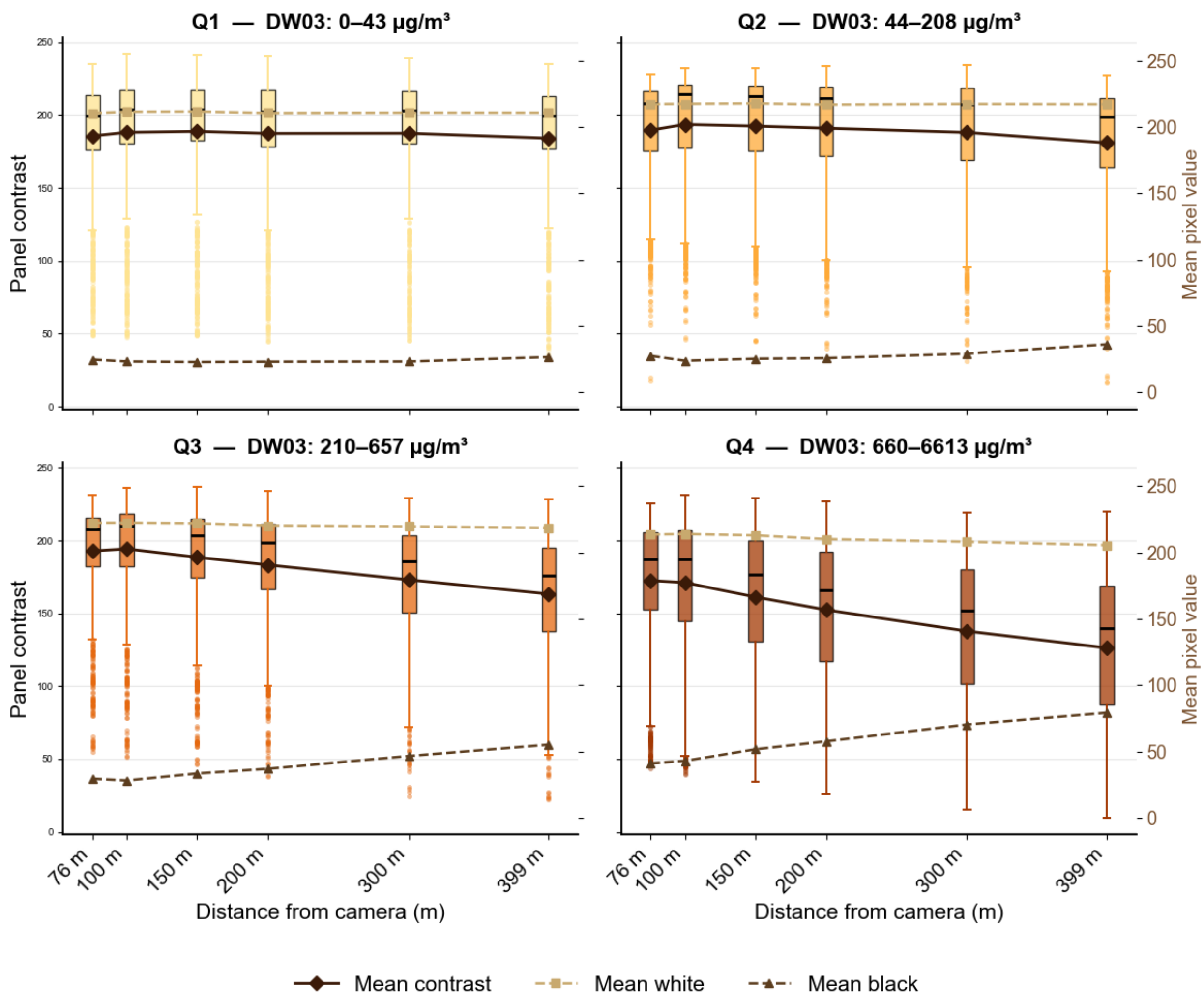

Figure 4. Panel contrast (white minus black pixel value) and mean pixel values of black and white panels as a function of distance from the camera, stratified by DW03 $PM_{10}$ concentration quartiles (Q1–Q4).

## 3.2 Machine Learning Model Performance and the Contribution of Temporal Lag Features and Solar Zenith Angle

Figure 5 presents preliminary 5-fold cross-validation results for three candidate models: RF, XGB, and QRF. XGB achieved the highest predictive accuracy with the highest $R^2$ and lowest NRMSE across both sensors, while QRF and RF performed comparably but slightly lower. QRF was retained as a candidate model despite its moderate accuracy due to its ability to provide prediction intervals, which is particularly valuable given the high RMSE levels observed in these preliminary results. Across all three models, prediction accuracy was consistently higher for DW03 than DW04, with DW04 exhibiting substantially larger RMSE values driven by extreme $PM_{10}$ concentrations reaching 10,000–20,000 μg $m^{-3}$. MedAE, by contrast, reflects a more moderate typical error of approximately 100–200 μg $m^{-3}$.Based on these preliminary results, XGB was selected for hyperparameter tuning and further evaluation as the primary prediction model.

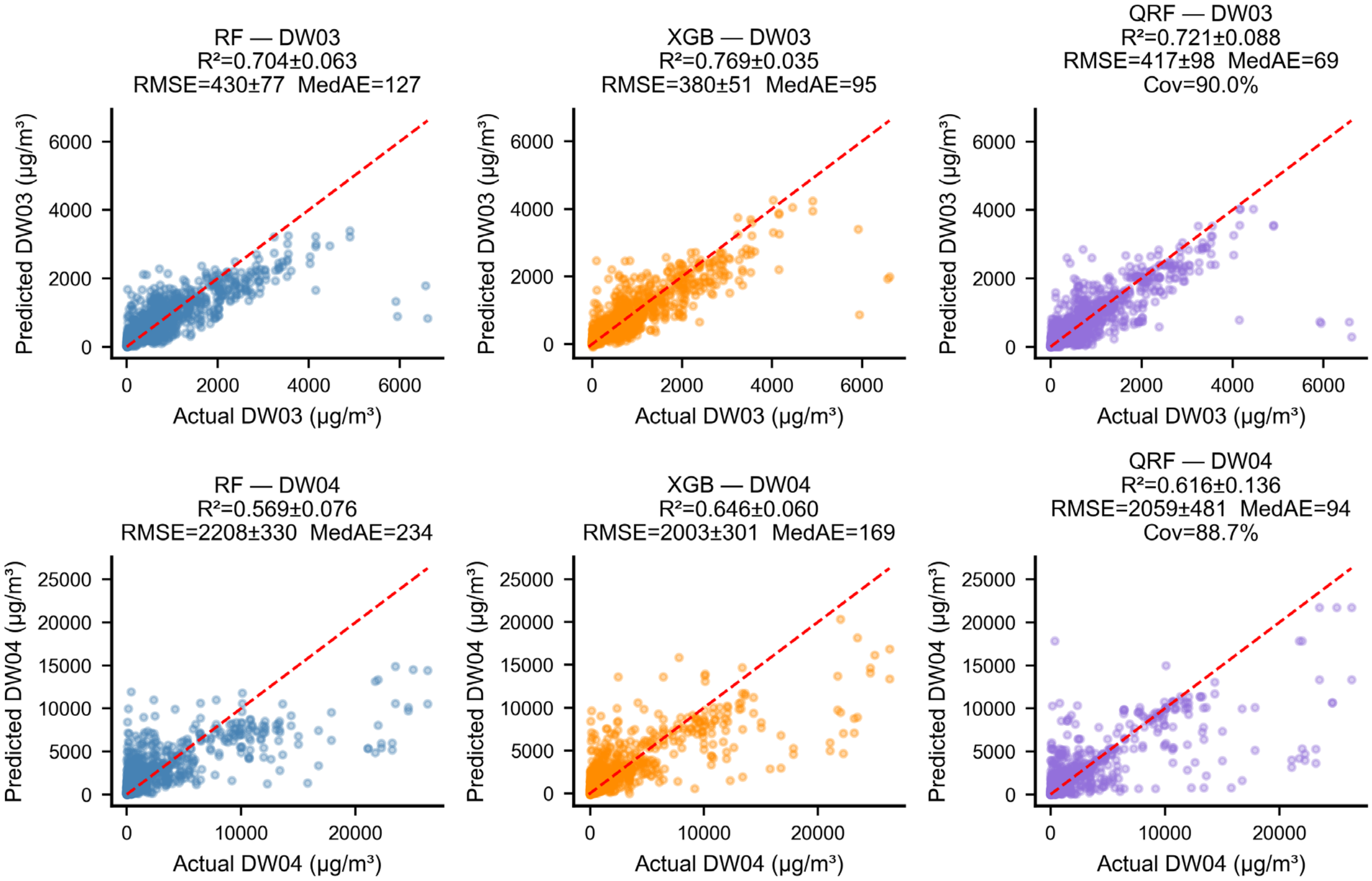

Fig 5. Predicted versus actual $PM_{10}$ concentrations from 5-fold cross-validation out-of-fold predictions for RFR, XGB, and QRF trained on image-derived features including panel contrast, black and white pixel values, their temporal lag values from the preceding five images, and solar zenith angle, to simultaneously predict DW03 and DW04. XGB achieved the highest accuracy for DW03, followed by QRF, and RFR. Prediction accuracy was consistently lower for DW04 across all models.

Figure 6 presents the effect of incrementally adding temporal lag features and solar zenith angle on XGB model performance. Lag features consist of contrast, black panel, white panel, and brightness values extracted from the previous n images captured on the same day. $R^2$ increased progressively for both sensors as the number of lag steps increased from zero to five, rising from 0.67 to 0.75 for DW03 and from 0.44 to 0.60 for DW04 (Figure 6a). RMSE for DW04 decreased steadily from approximately 2,500 µg $m^{-3}$ at no lag to approximately 2,200 µg $m^{-3}$ at Lag 5, while DW03 RMSE remained relatively stable throughout, around 480 µg $m^{-3}$ (Figure 6b). The addition of solar zenith angle at Lag 5 produced the largest single-step gain across both sensors, with $R^2$ increasing to 0.792 for DW03 and 0.677 for DW04. Relative to the no-lag baseline, the full Lag 5 + Zenith configuration improved $R^2$ by +0.124 and +0.238 for DW03 and DW04 respectively (Figure 6c), with corresponding RMSE reductions of 74 µg $m^{-3}$ and 588 µg $m^{-3}$ (Figure 6d).

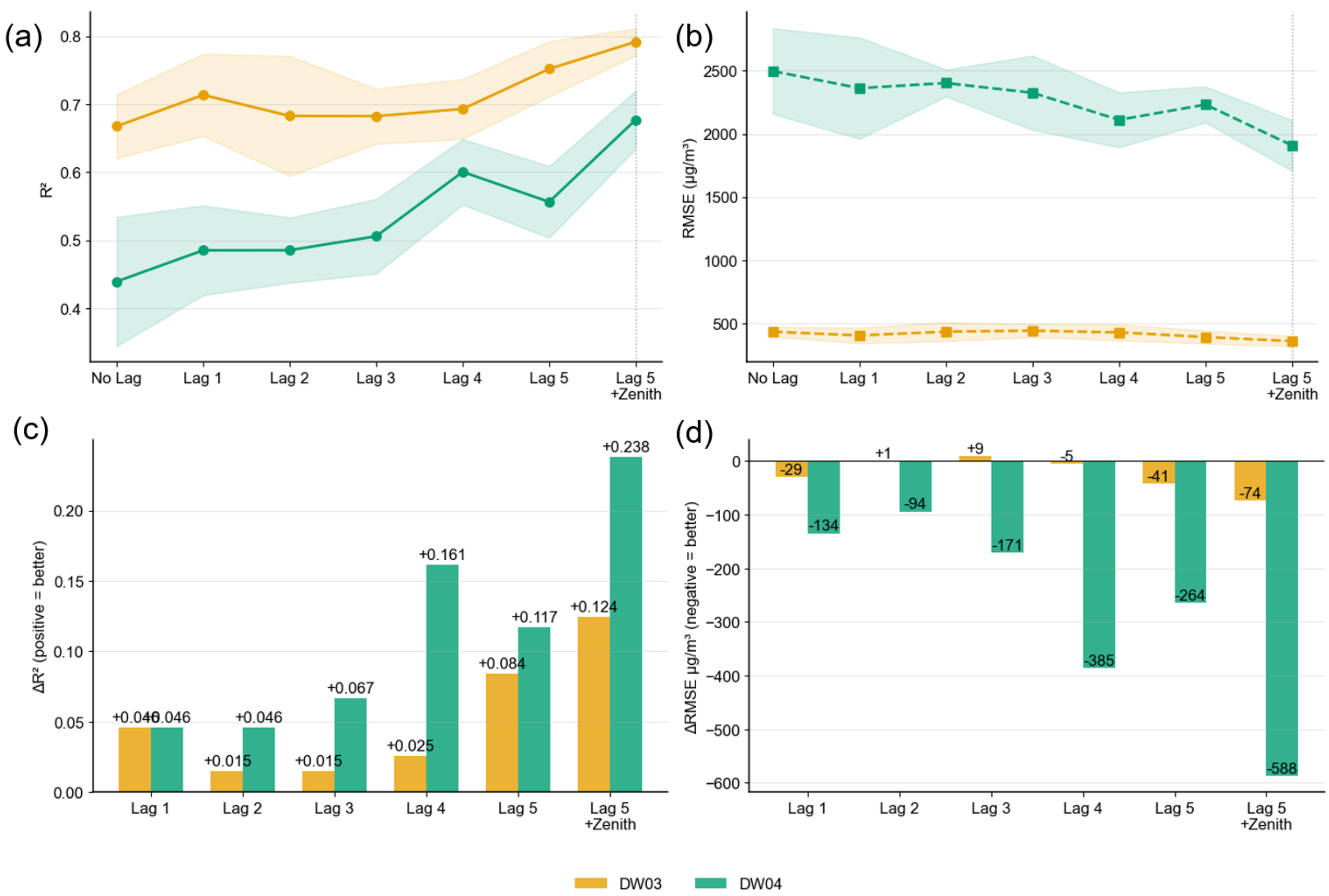

Fig 6. Effect of incorporating temporal lag features and solar zenith angle on XGB model performance, evaluated using 5-fold cross-validation. (a) $R^2$ and (b) RMSE as a function of the number of lag steps included, where each lag step represents one preceding image. (c) Absolute improvement in $R^2$ and (d) RMSE reduction relative to the no-lag baseline.

Figure 7 presents the final tuned XGB model performance and the top 25 most important features ranked by mean importance across both targets. The tuned model achieved $R^2$ of 0.792, NRMSE of 65.0%, and MedAE of 103 μg/m³ for DW03, and $R^2$ of 0.677, NRMSE of 133.1%, and MedAE of 194.3 μg/m³ for DW04 (Figure 7a). Among all features, the black panel pixel value of Panel 1, which is the panel located furthest from the camera, ranked as the single most important feature, followed by solar zenith angle as the second highest. Black panel pixel values from Panel 2 and Panel 4 also appeared prominently among the top features at the current timestep (t = 0), while panels closer to the camera such as panels 8 through 12 were rarely represented in the top 25. Notably, the raw black and white pixel values individually contributed more to model predictions than the derived contrast values themselves, with contrast features appearing only occasionally and exclusively as lagged terms. White panel lag features, particularly from Panels 2 and 12, also appeared among the top predictors despite the lower sensitivity of the white panel to $PM_{10}$ concentration changes reported in Figure 4a, suggesting that temporal changes in white panel values across previous images carry predictive signal not captured at the current timestep alone. Overall, the top 25 features are dominated by raw pixel values from panels at greater distances from the camera, temporal lag features, and solar zenith angle, with contrast features playing a secondary role.

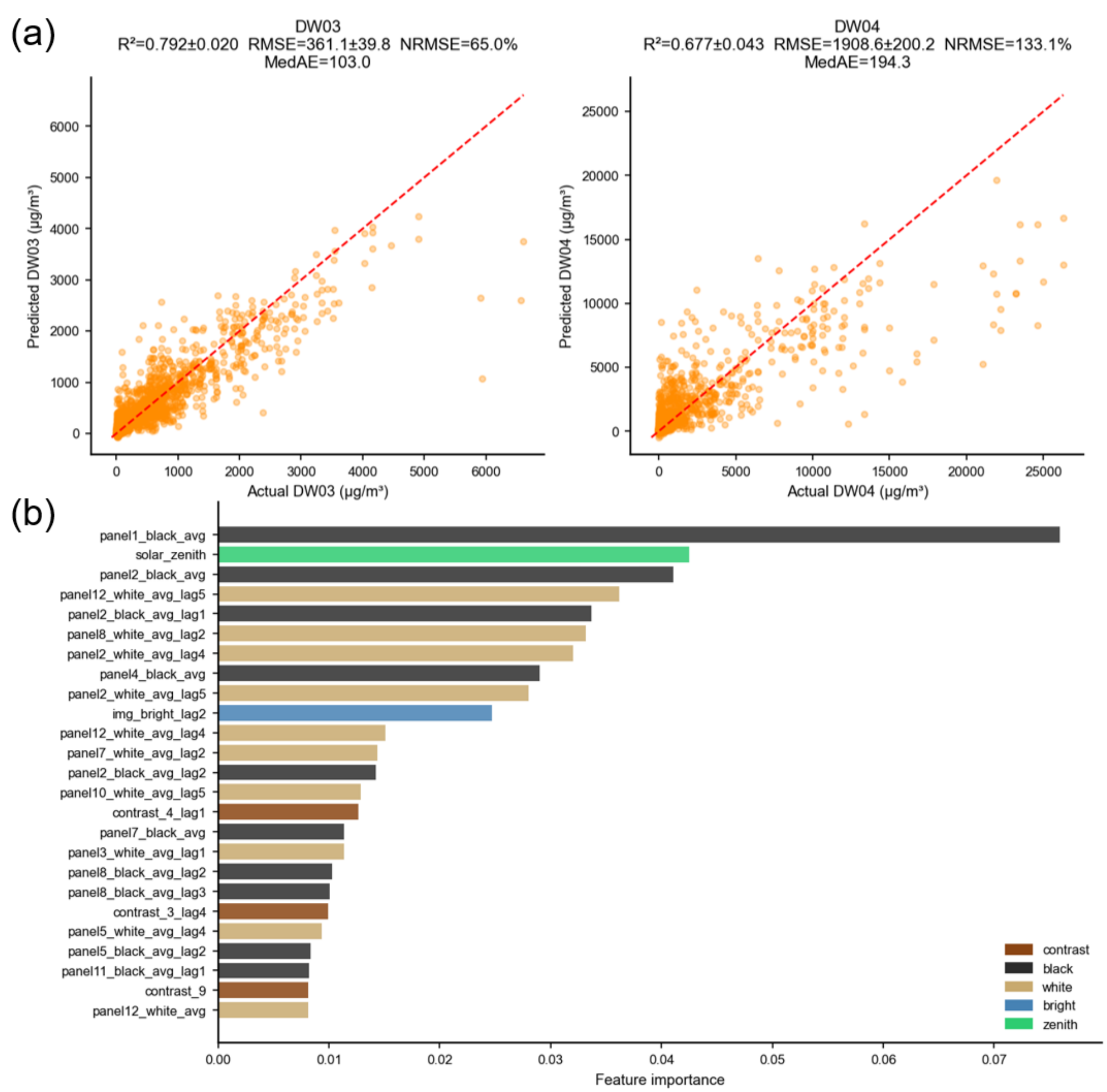

Fig 7. Performance of the tuned XGB model (n_estimators = 927, max_depth = 7, learning_rate = 0.22, subsample = 0.82) evaluated using 5-fold cross-validation. (a) Predicted versus actual $PM_{10}$ concentrations for DW03 and DW04 (b) Top 25 features ranked by mean importance across both targets.

Figure 8 presents the mean Laplacian variance at each panel position and distance from the camera, stratified by $PM_{10}$ concentration quartiles. Laplacian variance declined sharply beyond Panel 9, corresponding to distances lower than 200 m, which is attributable to the camera focus being calibrated to the 200 m target panels at the midpoint of the image. Panels 7 and 8 at 150 m retained relatively high Laplacian values, indicating acceptable image sharpness at that distance. Panels 1 through 6, which span distances of 200 m and beyond and were identified as high-importance features in the model, maintained acceptable sharpness levels within the camera's effective focal range. Although Laplacian variance showed a statistically significant inverse relationship with $PM_{10}$ concentration, with a Pearson correlation of $r = -0.887$ ($p = 0.045$) between Panel 1 Laplacian variance and DW03, incorporating Laplacian variance as a model feature did not improve prediction performance in preliminary testing and was therefore excluded from the final feature set.

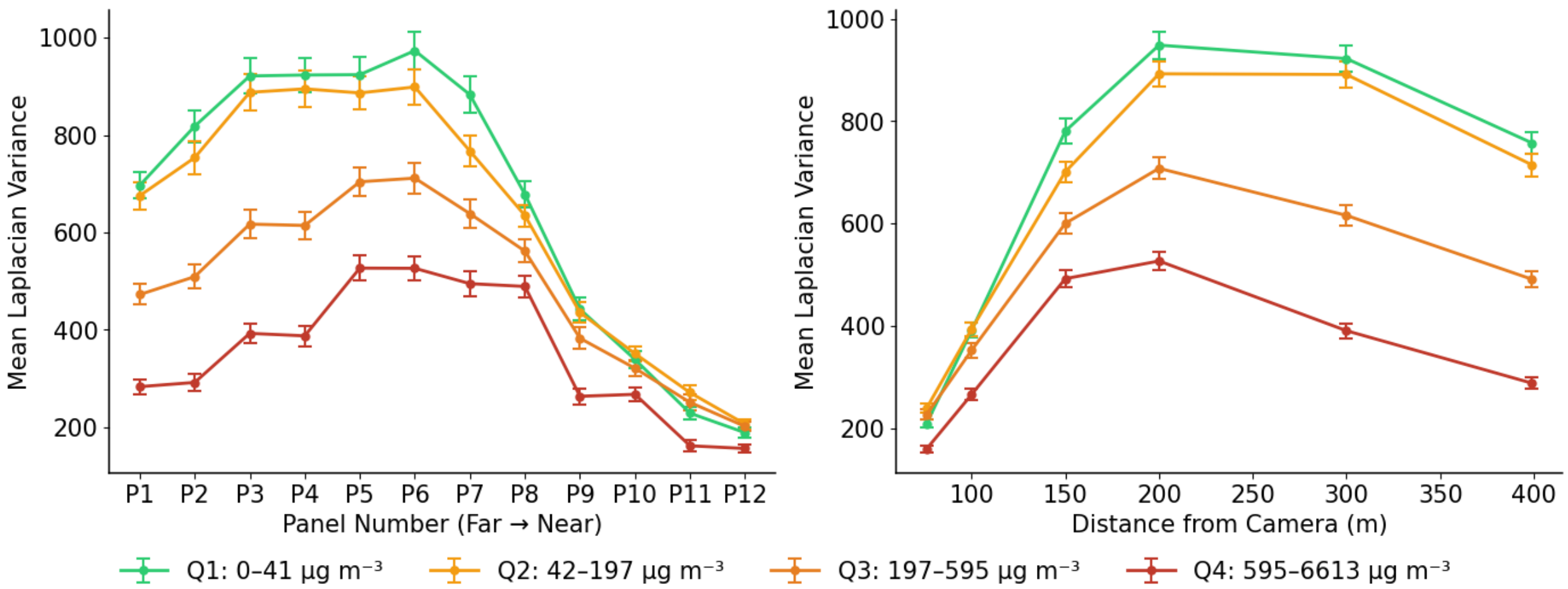

Figure 8. Mean Laplacian variance of each panel region as a function of panel number and distance from the camera, stratified by DW03 $PM_{10}$ concentration quartiles. Error bars represent the 95% confidence interval of the mean.

Figure 9 illustrates the progressive growth in residual magnitude during the sunset transition, with errors reaching their maximum as the solar zenith angle approaches 90 degrees. The smaller residuals observed at DW03 relative to DW04 are attributable to the substantially lower $PM_{10}$ concentrations recorded at DW03, as shown in Figure 2. Residuals prior to 18:00 remain low across both sensors, although this likely reflects the limited dynamic range of $PM_{10}$ concentrations during the pre-peak period rather than superior model accuracy under daylight conditions. Taken together, these patterns indicate that model uncertainty is concentrated within the sunset window, when rapidly changing illumination and the onset of the evening dust peak coincide.

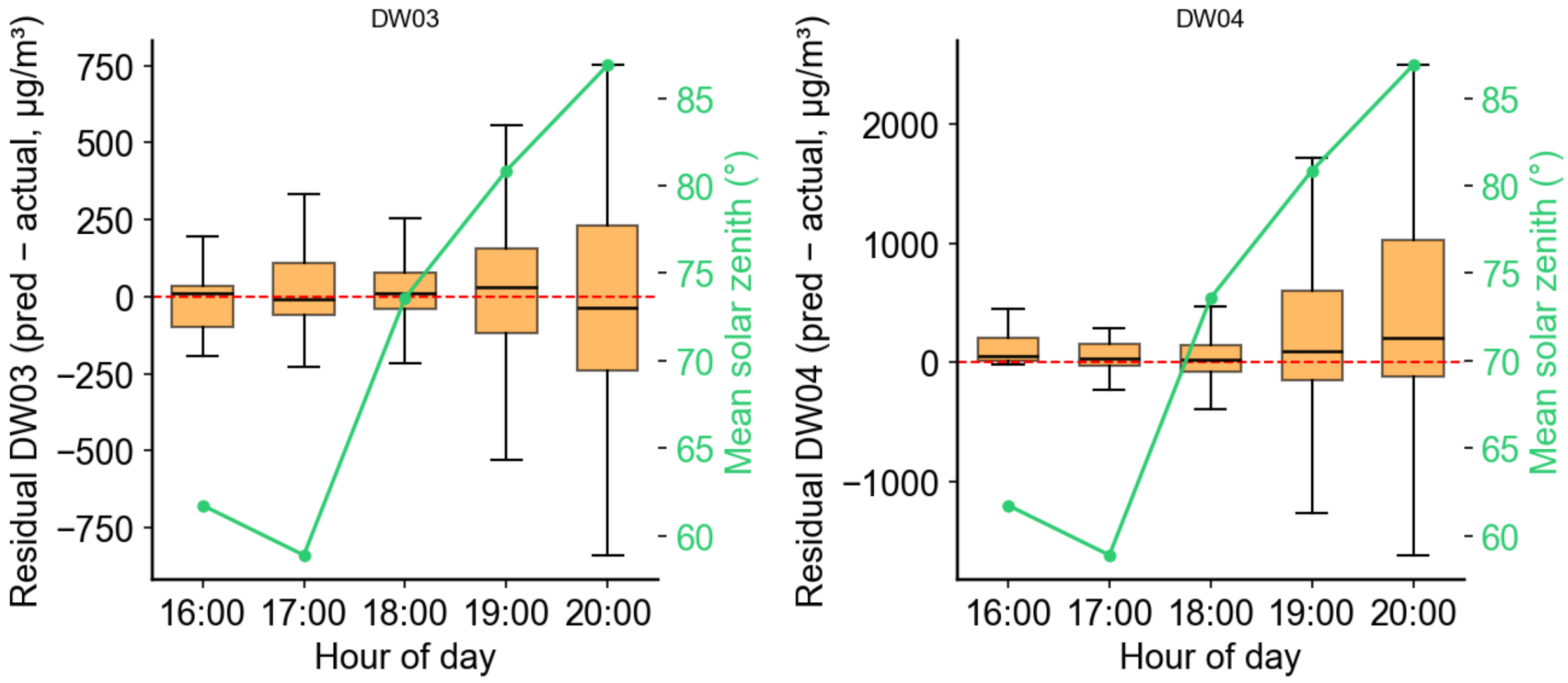

Figure 9. Hourly distribution of prediction residuals (predicted minus actual $PM_{10}$) for DW03 and DW04, overlaid with mean solar zenith angle. Residual spread widens markedly as solar zenith angle approaches 90 degrees near sunset.

# 4. Discussion

## 4.1 Model Accuracy, Feature Contribution, and Sources of Prediction Uncertainty

The incorporation of temporal lag features and solar zenith angle produced consistent improvements in model performance across both sensors, as demonstrated in Figures 6 and 7. The progressive increase in $R^2$ and reduction in RMSE with each additional lag step confirms that short-term temporal context carries meaningful predictive information, a finding further supported by the appearance of lagged pixel values among the top ten most important features in the final model. Solar zenith angle ranked as the second most important feature overall, indicating that it effectively encodes ambient lighting conditions that directly influence panel pixel values. In contrast, whole-image brightness and panel brightness contributed negligible importance and could be excluded from future model configurations without meaningful loss of predictive accuracy.

DW03 consistently outperformed DW04 across all model configurations. This disparity may reflect the spatial relationship between each sensor and the camera system. As the nearer downwind sensor, DW03 monitors an air column that intersects the optical path of all twelve panel pairs, meaning that particulate matter in this column contributes to contrast reduction across the entire image. DW04, positioned further downwind, monitors an air column that corresponds to only the most distant panel pairs (panels 1 through 4). This geometric arrangement is consistent with observations from optical remote sensing studies of volcanic plumes, where increasing distance between the camera and the target plume has been shown to degrade instrument sensitivity to column optical depth (Kern et al., 2013), and suggests that the near-field air column captured by DW03 is inherently better represented by the image-derived contrast signal.

Although $R^2$ values of 0.792 and 0.677 for DW03 and DW04 respectively indicate reasonable predictive skill, NRMSE values of 65.0% and 133.1% are substantially higher than those reported in comparable urban studies, including Wang et al. (2024) at approximately 25% and Liu et al. (2024) at approximately 45%, though it should be noted that Lit et al. (2024) studies targeted $PM_{2.5}$, a particulate fraction more compellingly associated with light extinction.

The elevated NRMSE in the present study is largely attributable to the extreme $PM_{10}$ concentrations encountered in feedlot environments. DW04 recorded peak 5-minute averaged concentrations approaching 20,000 μg $m^{-3}$ during high dust events, and these infrequent but extreme observations disproportionately inflate RMSE due to the squared error penalty inherent in its calculation, which in turn drives NRMSE upward when normalized against the overall mean concentration. This sensitivity of RMSE and NRMSE to extreme values is a well-recognized limitation of these metrics when applied to data set with outliers (Hassanat et al., 2024; Mundu et al., 2026). The comparatively lower NRMSE for DW03 further supports this interpretation, as DW03 recorded fewer extreme concentration events, resulting in a less skewed error distribution despite operating under the same model and feature set. In this context, the MedAE offers a more robust performance measure, as it is largely unaffected by extreme observations and reflects the typical prediction error rather than being dominated by rare high-concentration events.

## 4.2 Temporal Overlap Between Sunset Illumination Decline and the Evening Dust Peak

Although the inclusion of solar zenith angle provided the model with contextual information regarding incoming illumination and improved predictive accuracy (Figure 7), residual analysis indicates that considerable challenges remain during the sunset transition (Figure 8). The largest prediction errors were consistently observed during the period in which ambient light intensity diminishes most rapidly, which coincides with the onset of the evening dust peak. This diurnal pattern of elevated $PM_{10}$ emissions during evening hours has been well documented in cattle feedlot environments and is generally attributed to declining solar radiation, decreasing ambient temperature, and intensified animal activity (Auvermann et al., 2010; Hiranuma et al., 2011; McGinn et al., 2010; Peanusaha et al., 2026). The temporal coincidence of these three phenomena presents a fundamental challenge for image-based estimation, as the period exhibiting the greatest $PM_{10}$ dynamic range is also the period in which illumination conditions are most variable, thereby compounding the optical signal attributable to dust with that attributable to changing ambient light.

This limitation is not unique to feedlot applications. Comparable degradation in model performance under low-light or nighttime conditions has been reported in urban image-based air quality studies, where reduced illumination similarly compromises the extraction of dust-related visual features (Kow et al., 2022; Song et al., 2020; Xiang et al., 2025). To address this constraint, the integration of supplemental artificial illumination, as employed in several urban studies, represents a promising direction for feedlot deployments. However, the design of such a system must account for the potential of artificial light sources to induce image saturation. A practical implementation would likely require a gradually intensifying artificial light source synchronized with the declining availability of natural light, thereby preserving panel contrast across the full sunset transition without introducing abrupt changes in image radiometry. Future work should evaluate the calibration requirements and operational feasibility of such a hybrid illumination system in field conditions.

## 4.3 Deployment Guidelines for Panel Placement, TEOM Reference Station Positioning, and Blur Prevention

The Laplacian score distribution and feature importance analysis together provide practical guidance for panel placement in operational deployments. Panels positioned closer than the camera focal length, particularly Panel 9 and beyond, exhibited lower Laplacian scores indicative of focus-induced blur (Figure 8), and panels beyond Panel 6 contributed negligibly to model predictions (Figure 7). These results suggest that future deployments should prioritize panel placement at intermediate to far distances within the camera depth of field, rather than maximizing panel count through near-field positioning. With respect to reference instrument placement, the TEOM monitor should be positioned at the centerline of the dominant plume axis to capture representative $PM_{10}$ concentrations during peak emission periods, consistent with the higher concentrations recorded at DW04. Equally important, the reference instrument should be located close to, or immediately in front of, the panel array, such that the air column sampled by the TEOM is representative of the air column through which light reflected from the panels must travel before reaching the camera.

These findings are also consistent with the Gaussian plume assumption described by Upadhyay (2008), in which the highest $PM_{10}$ concentrations are expected at the middle point of the plume cross-section relative to the prevailing wind direction. The consistently higher concentrations recorded at DW04 compared to DW03 support this interpretation, as DW04 is positioned closer to the estimated plume centerline under the predominantly southerly wind conditions observed during the study period (Appendix 2). A practical implication of this geometric relationship is that the predictive accuracy of the image-based system for any given reference point is maximized when the camera is positioned such that the air column of interest lies directly between the camera and the panel array, thereby ensuring that contrast reduction captured in the image is representative of the $PM_{10}$ concentration at the target location. In operational terms, if the monitoring objective is to estimate peak plume concentrations at a location equivalent to DW04, the camera station should be co-located with or positioned immediately upwind of that reference instrument, rather than offset laterally as in the present configuration.

## 4.4 Remaining Challenges and Future Work

A persistent challenge remains in predicting $PM_{10}$ during the sunset transition, when rapid changes in ambient illumination coincide with the onset of the evening dust peak. Future work should investigate the deployment of supplemental artificial lighting, calibrated to gradually compensate for the declining natural light, as a strategy to mitigate this confounding effect and further improve model reliability during the critical evening dust event.

# Acknowledgement

This research was supported by USDA-NIFA Award No. 2009-55112-053235, Lamar University, Texas Cattle Feeder Association (TCFA), and Texas A&M AgriLife Research.

# Credit authorship contribution statement

**Sirapoom Peanusaha** –Formal analysis; Data curation; Software; Investigation; Methodology; Supervision; Writing—original draft; Writing—reviewing & editing.

**Greg Ferguson**— Project administration; Data curation; Resource; Writing—original draft; Writing—reviewing & editing.

**K. Jack Bush**— Project administration; Writing—reviewing & editing

**Peiyang Li**—Validation; Writing—reviewing & editing.

**Brent W. Auvermann** – Conceptualization; Resource; Funding acquisition; Supervision; Project administration; Writing—reviewing & editing.

# Appendix

Appendix 1. Comparison of $PM_{10}$ Monitoring Instruments and Camera System by Measurement Capability and Cost.

| Instrument | Primary Use | $PM_{10}$ Measurement Range (μg m⁻³) | Accuracy[1] | Approximate Cost |
|---|---|---|---|---|
| Davis AirLink 7210 | Indoor/outdoor consumer air quality | 2.5–10 μm | ±10 μg $m^{-3}$ | $180–$225 |
| Temtop PMD 351 | Indoor/handheld spot checks | 0–1,000 | ±10% | $500–$800 |
| Met One ES-642 | Outdoor industrial monitoring | 0–1,000 | ±5% | $6,500–$8,500 |
| Thermo Fisher 1405 TEOM | Regulatory ambient monitoring | 0–1,000,000 | ±0.75% | $30,000–$35,000 |
| DusTrak DRX Aerosol 8533 | Desktop model for area and process monitoring | 0.001 – 150,000 | ±0.1% | $14,000 - $18,000 |
| Canon EOS 60D DSLR | Image capture (this study) | N/A | N/A | ~$700 |

[1] based on manufacturer's specification

Appendix 2. Wind roses showing (a) the full diurnal wind profile at the study site based on all available data from 2023 to 2025 and (b) the wind profile restricted to evening image acquisition periods, showing predominantly southerly wind direction consistent with the positioning of downwind sensors DW03 and DW04.

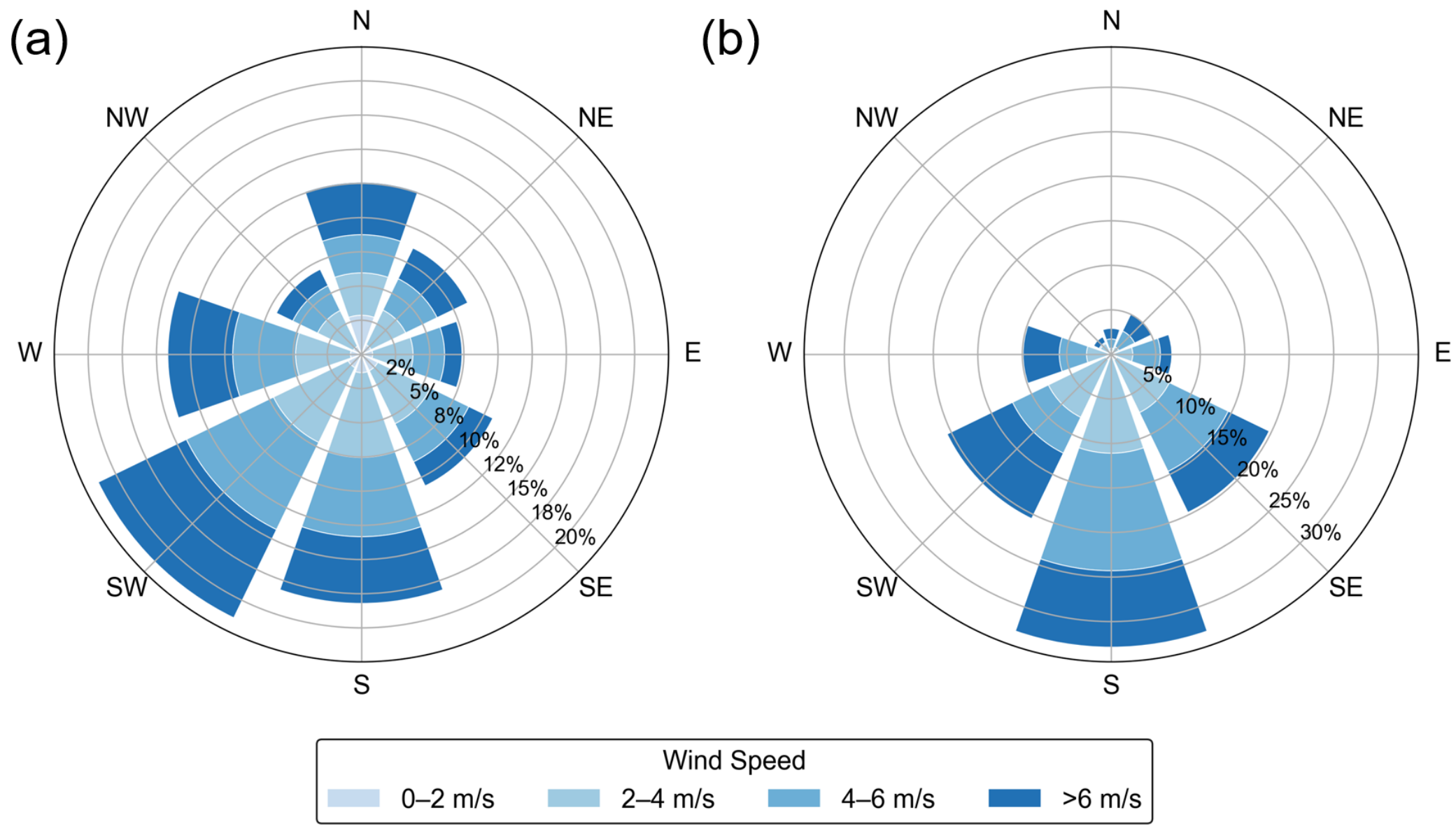

Appendix 3. Schematic of the initial panel segmentation algorithm, illustrating how paired black and white targets are detected and how their centroids are referenced along the horizontal (x) and vertical (y) image axes.

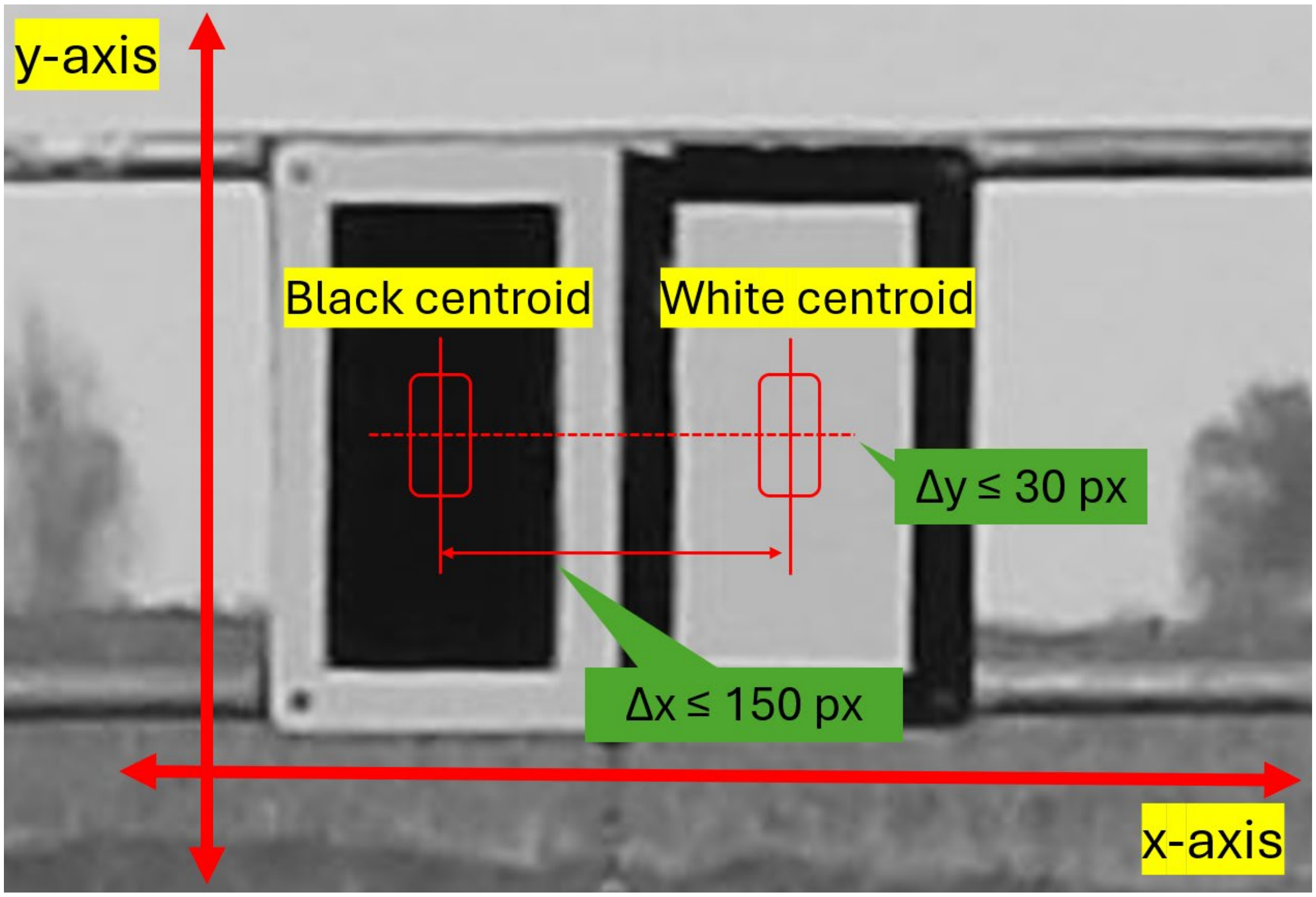